\documentclass{styles/svproc}
\usepackage{amsmath}
\usepackage{amssymb}
\usepackage{float}
\usepackage{booktabs}
\usepackage{hyperref}
\usepackage{algorithm}
\usepackage{algorithmicx}
\usepackage{mathtools}
\usepackage{algpseudocode}
\usepackage{listings}
\usepackage{xcolor}

\DeclareMathOperator*{\argmin}{arg\,min}

\begin{document}
\mainmatter 
\title{Sparse clustering via the Deterministic Information Bottleneck algorithm}
\titlerunning{Sparse clustering via the Deterministic Information Bottleneck algorithm} 
\author{Efthymios Costa\inst{1} \and Ioanna Papatsouma\inst{1}
\and Angelos Markos\inst{2}}
\authorrunning{Efthymios Costa et al.}
\tocauthor{Efthymios Costa, Ioanna Papatsouma, and Angelos Markos}
\institute{Department of Mathematics, Imperial College London, SW7 2AZ, United Kingdom,\\
\email{efthymios.costa17@imperial.ac.uk}, 
\email{i.papatsouma@imperial.ac.uk},
\and
School of Education, Department of Primary Education, Democritus University of Thrace, GR-68132, Nea Hili, Alexandroupoli, \\
\email{amarkos@eled.duth.gr}}

\maketitle

\begin{abstract}
Cluster analysis relates to the task of assigning objects into groups which ideally present some desirable characteristics. When a cluster structure is confined to a subset of the feature space, traditional clustering techniques face unprecedented challenges. We present an information theoretic framework that overcomes the problems associated with sparse data, allowing for joint feature weighting and clustering. Our proposal constitutes a competitive alternative to existing clustering algorithms for sparse data, as demonstrated through simulations on synthetic data. The effectiveness of our method is established by an application on a real-world genomics data set.

\end{abstract}

\section{Introduction}\label{sec:intro}

Data exhibiting feature-level sparsity, where the relevant signal resides in a small subset of variables, is nowadays commonly encountered in several research areas. For instance in bioinformatics, gene expression data is inherently high-dimensional and sparse, or in chemometrics, the analysis of spectra for material characterisation or product quality control typically involves very few samples and a large number of features. Clustering algorithms can be very useful for assigning objects into groups based on their similarities. However, they are typically accounting for all the variables included in a data set, thus undermining the interpretability of the obtained partition when just a subset of the features is informative about an existing lower-dimensional cluster structure. This inclusion of uninformative variables can obscure the underlying signal and mislead the clustering process, resulting in incorrect partitions. Moreover, distance-based clustering algorithms suffer from the curse of dimensionality for large enough numbers of variables, whereas model-based techniques struggle with singularity issues when the number of features exceeds that of samples.

In this paper, we present an information-theoretic algorithm for clustering sparse and potentially high-dimensional data. The proposed approach is based on the Information Bottleneck algorithm (IB) \cite{tishby99information} and extends its application to cluster analysis that was presented in \cite{costa2026deterministic}. The rest of the paper is organised as follows; in Section \ref{sec:dibclust} we present the deterministic variant of IB and how it can be used for clustering. Section \ref{sec:sparsedib} extends the clustering framework to account for feature-level sparsity, allowing for feature weighting to be performed simultaneously with the clustering. Simulations on synthetic data are presented in Section \ref{sec:simstudy}, where the proposed method is compared to six algorithms specifically designed for sparse clustering. We illustrate the applicability of our approach on a genomics data set in Section \ref{sec:applications}. We finally give some concluding remarks and future study recommendations in Section \ref{sec:conclusion}.

\section{Deterministic Information Bottleneck Clustering}\label{sec:dibclust}

Traditional clustering methods rely on geometric assumptions or arbitrary distance metrics. Information-theoretic alternatives frame clustering as an optimisation problem, seeking compressed representations that retain maximal information. A characterisation of this compression-relevance tradeoff that produces a crisp partition can be obtained using the deterministic variant of the Information Bottleneck algorithm (DIB) \cite{strouse2017deterministic}. This Section summarises these ideas for continuous data.

In the clustering paradigm, the random variables $Y, T,$ and $X$ are used to model the observed feature values, the cluster assignment, and the observation index (treated as uninformative). The DIB algorithm assumes that $T$ and $Y$ are conditionally independent given $X$ and seeks a deterministic encoder $q^*(t \mid x)$ satisfying:
\begin{equation}\label{eq:dib}
    q^*(t \mid x) = \argmin\limits_{q(t \mid x)} H(T) - \beta I(Y; T),
\end{equation}
where $H(T)$ is the entropy of $T$ (cluster compression measure), $I(Y;T)$ the mutual information (cluster relevance measure), and $\beta > 0$ controls the compression-relevance tradeoff. Solving this optimisation problem yields a partition where each cluster is defined as a group of points whose feature distributions are well approximated by a common prototype $q(\boldsymbol{y} \mid t)$, with assignments driven by information preservation rather than geometric distance. Following \cite{costa2026deterministic}, we search for a partition maximising $I(Y;T)$ with $\beta$ updated dynamically to prevent empty clusters, essentially treating $H(T)$ as a regularisation term for cluster imbalance.

The core of DIB is the perturbed similarity matrix $\mathbf{P}'_{Y \mid X} \in \mathbb{R}^{n \times n}$, with entries $[\mathbf{P}'_{Y \mid X}]_{i,j} = p(Y = \boldsymbol{y}_i \mid X = x_j)$ representing probabilistic proximities. Density estimation uses product kernels \cite{li2003nonparametric}:
\begin{equation*}
    p(Y = \boldsymbol{y}_i \mid X = x_j ) \propto \prod_{m=1}^p K_m(y_{i,m}, y_{j, m}; \lambda_m),
\end{equation*}
where $K_m(\cdot, \cdot; \lambda_m)$ is a kernel function (we use the Gaussian kernel) with bandwidth $\lambda_m$, chosen via the criterion from \cite{costa2026deterministic} or set manually. The algorithm iteratively updates cluster distributions and allocates observations to clusters maximising $\log[q(t)] - \beta \text{D}_\text{KL}[ p(\boldsymbol{y}_i \mid x) \lvert \rvert q(\boldsymbol{y}_i \mid t)]$, where $\text{D}_\text{KL}(\cdot \lvert \rvert \cdot)$ is the Kullback-Leibler (KL) divergence. The term $q(t) = \sum_j q(t \mid x_j)p(x_j)$ is the mass function of the random variable describing the cluster masses, whereas $q(\boldsymbol{y}_i \mid t) = \sum_j q(t \mid x_j)p(\boldsymbol{y}_i \mid x_j)p(x_j)/q(t)$ quantifies the expected feature distribution of points in cluster $t$. The latter is optimised when the cluster-conditional density estimator matches the feature distribution of points in that cluster, rendering a higher cluster relevance term $I(Y;T)$, or equivalently, a lower KL divergence. Multiple random initialisations are used and the solution yielding the highest mutual information value is selected.
\vspace{-0.1cm}
\section{An extension for sparse clustering}\label{sec:sparsedib}

In sparse clustering, only a small proportion of variables is informative about the underlying cluster structure. Standard techniques like K-Means or DIB assume all variables are equally informative, which is rarely true in practice. We extend DIB to incorporate feature weighting; feature selection is a special case, with zero weights corresponding to unselected variables.

The Sparse DIB optimisation problem becomes:
\begin{align}\label{eq:sparsedib}
    q_W^*(t \mid x) = \argmin\limits_{q_W(t \mid x), \mathbf{w}} H(T) - \beta I(Y_W; T),\\ \text{subject to } \|\mathbf{w} \|_2 \leq 1, \ \| \mathbf{w} \|_1 \leq u, \ w_j \geq 0 \ \forall 1\leq j \leq p, \nonumber
\end{align}
where $Y_W$ is the weighted random variable and $\mathbf{w} \in \mathbb{R}^p$ are feature weights constrained to the first orthant of the unit $L_2$-ball with an $L_1$ constraint controlled by sparsity parameter $u > 0$. The weighted perturbed similarity matrix is:
\begin{equation*}
    \left[\mathbf{P}^{'W}_{Y \mid X}\right]_{i,j} \propto \prod_{m=1}^p K_m(y_{i,m}, y_{j,m}; \lambda_m)^{w_m},
\end{equation*}
with weights introduced exponentially to control each variable's contribution. Note that for some kernel functions such as the Gaussian, this reduces to bandwidth rescaling ($\lambda_m \leftarrow 
\lambda_m/\sqrt{w_m}$) but we keep the exponentiated form for broader applicability. The algorithm (summarised in Algorithm \ref{code:sparsedib}) alternates between obtaining cluster assignments via DIB and updating weights until these converge (we set a convergence tolerance threshold $\epsilon = 10^{-5}$). Weights are updated using Dykstra's projection algorithm \cite{dykstra1985iterative} by projecting onto the feasible set $\mathcal{C} = \{ \mathbf{w}: \|\mathbf{w} \|_2 \leq 1, \| \mathbf{w} \|_1 \leq u, \mathbf{w} > \mathbf{0}\}$, giving us a vector of weights $\mathbf{w}$ that satisfies both $L_1$ and $L_2$ constraints.

\begin{algorithm}[!ht]
\caption{Sparse DIB clustering}
\label{code:sparsedib}
\begin{algorithmic}[1]
\State \textbf{Input:} Data $\mathbf{X} \in \mathbb{R}^{n \times p}$, sparsity constraint $u$, max iterations $m^\text{max}$, threshold $\epsilon$
\State \textbf{Output:} Cluster assignment $q^*_W(t \mid x)$, weights $\mathbf{w}^*$
\Statex
\State Initialise $\mathbf{w}^{(0)}$ (uniform or warm start) and set $m \leftarrow 1$
\While{not converged and $m \leq m^\text{max}$}
  \State Obtain $q^{(m)}_W(t \mid x)$ using DIB with weights $\mathbf{w}^{(m-1)}$
  \State Update $\mathbf{w}^{(m)}$ by setting $w_j \propto I(Y_j; T)$ using Dykstra's algorithm
  \State Check convergence; $\sum_j |w_j^{(m)} - w_j^{(m-1)}| / \sum_j |w_j^{(m-1)}| < \epsilon$ and set $m \leftarrow m+1$
\EndWhile
\end{algorithmic}
\end{algorithm}

Weights can be initialised uniformly ($1/\sqrt{p}$) or via warm start from an initial K-Means solution. To tune $u$, we run the algorithm across multiple values and plot the normalised entropy of weights against $u$; a plateau indicates the range of suitable values for which the same features are selected as informative.

\section{Simulation Study}\label{sec:simstudy}

We conduct a simulation study in which Sparse DIB is benchmarked against six alternative clustering methods for sparse data, namely Sparse K-Means \cite{witten2010framework}, Random Projection Ensemble Clustering (RPEClust)\cite{anderlucci2022high}, model-based clustering with variable selection (VarSelLCM) \cite{marbac2017variable}, Clustering On Subsets of Attributes followed by Partitioning Around Medoids (COSA/PAM) \cite{friedman2004clustering,kaufmanrousseeuw1990} and Principal Component Analysis (PCA) and its sparse extension \cite{zou2006sparse}, followed by K-Means. We experimented with several hyperparameter values for each method and selected solutions that maximised the ARI. Uniform initialisation was used for Sparse DIB. Code for reproducing the simulations is publicly available in \url{https://github.com/EfthymiosCosta/SparseDIB_Sims}.

We generated synthetic data following a Gaussian mixture model on $\rho = \lfloor p \times \mathrm{q} \rfloor$ informative features ($\mathrm{q}$ being the proportion of informative features), with the remaining $p-\rho$ features drawn independently from standard Gaussians. We fixed $n = 200$ and varied $p \in \{100, 200, 400, 1000\}$, $\mathrm{q} \in \{0.05, 0.10, 0.20, 0.50\}$, and $K \in \{3, 5, 8\}$, with balanced/unbalanced and spherical/elliptical configurations (192 settings, 50 replicates each). Performance was assessed using the Adjusted Rand Index (ARI) \cite{hubert1985} and Adjusted Mutual Information (AMI) \cite{vinh_information_2010}.

\begin{figure}[!ht]
    \centering
    \includegraphics[width=\linewidth]{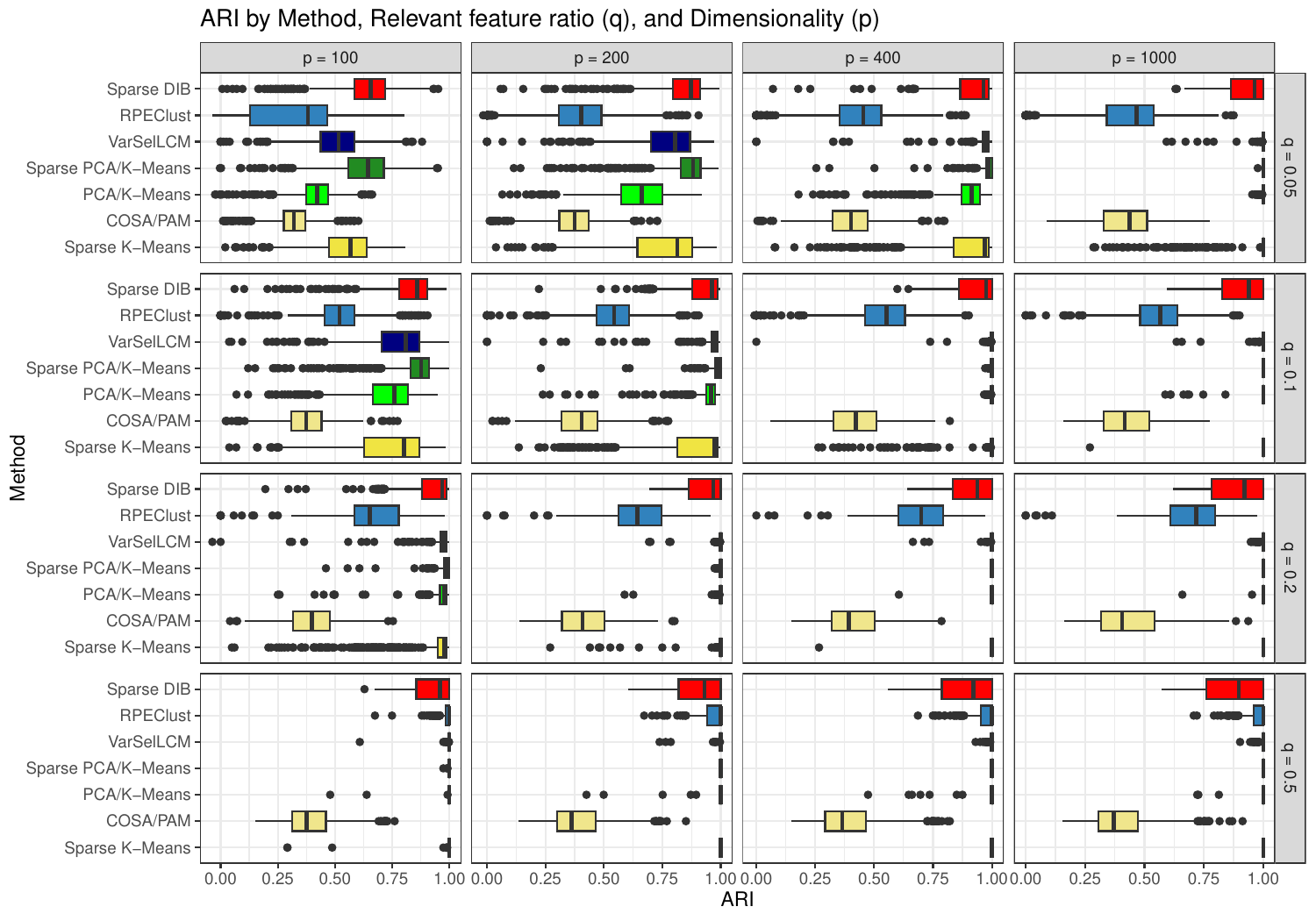}
    \caption{Box plots of ARI values across different dimensionalities ($p$) and ratios of relevant features ($\mathrm{q}$) for each method.}
    \label{fig:ari_RR_dim}
\end{figure}

According to Figure \ref{fig:ari_RR_dim}, the most successful methods in recovering cluster structure are VarSelLCM and dimension reduction approaches, with Sparse PCA performing best overall. Sparse DIB performs similarly to Sparse K-Means (mean ARI/AMI: 0.88/0.89 vs 0.91/0.92), highlighting their methodological similarities. COSA/PAM and RPEClust perform suboptimally, with the former never achieving perfect recovery. Notably, Sparse DIB outperforms competitors when $\rho$ is very small (e.g., $p = 100$, $\mathrm{q} = 0.05$); as $\mathrm{q}$ increases, most methods converge to near-perfect recovery. We omit the results for $K$, cluster imbalance, and sphericity for brevity, as performance variations were dominated by changes in $p$ and $\mathrm{q}$.

\begin{figure}[!ht]
    \centering
    \includegraphics[width=0.87\linewidth]{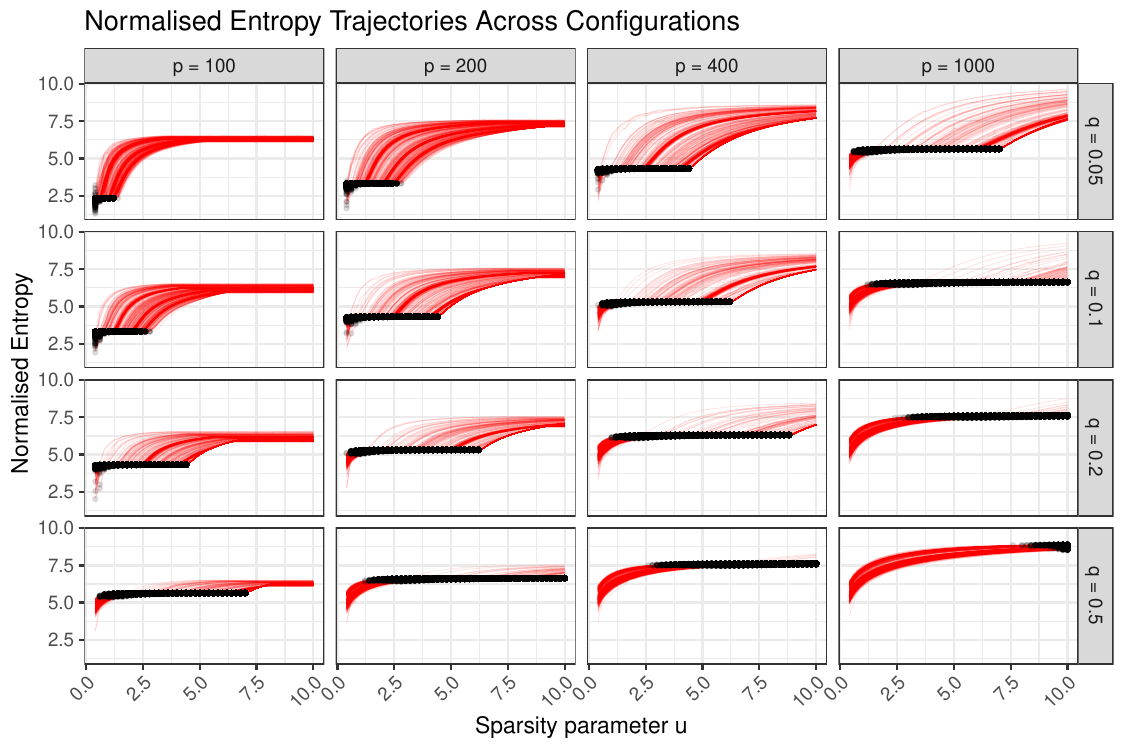}
    \caption{Trajectories of normalised entropy for sparsity parameter $u \in [0.4, 10]$. The black points mark the value of $u$ and the corresponding normalised entropy where the number of non-zero weights is closest to $\rho = \lfloor p \times \mathrm{q} \rfloor$.}
    \label{fig:entropy_trajectories_plots}
\end{figure}

We finally investigate the effect of the sparsity constraint value $u$ to the performance of Sparse DIB. Recalling that a cluster structure was deliberately defined on a fixed number of variables, we assess whether these features are correctly assigned non-zero weights. Figure \ref{fig:entropy_trajectories_plots} shows that the heuristic method described in Section \ref{sec:sparsedib} identifies the true number of relevant variables in most scenarios. The only exception is probably $p = 1000$ and $\mathrm{q} = 0.5$, where values of $u$ greater than ten would be needed for a clearer picture, hinting that a large number of simulations may be needed to be confident about the choice of $u$.
\vspace{-0.5cm}

\section{Application on Bladder Cancer Data}\label{sec:applications}

Biomedical data are often sparse and high-dimensional, particularly in cancer genomics where molecular features far outnumber available samples. To demonstrate our approach under such conditions, we analysed RNA-seq gene expression data from The Cancer Genome Atlas (TCGA) bladder cancer cohort (BLCA), accessed via the Genomic Data Commons portal \cite{cancer2014comprehensive}. Expression values (transcripts per million, TPM) were $\log_2(\text{TPM}+1)$ transformed. We retained samples with molecular subtype annotations and filtered to protein-coding genes with variance above 0.01, yielding 412 samples and 18193 genes. The five molecular subtypes were aggregated into three classes: Basal (144 samples), Luminal (248 samples), and Neuronal (20 samples). Sparse DIB was run using a warm start (K-Means with 1000 random initialisations) for sparsity values $u$ from 0.1 to 100. Results are summarised in Table \ref{tab:bladder_cancer_res}.

\begin{table}[!ht]
\centering
\setlength{\tabcolsep}{10pt}
\caption{ARI scores and number of selected features for sparse clustering algorithms applied to the cleaned bladder cancer data set.}
\begin{tabular}{lcc}
\toprule
Algorithm & ARI & Selected Features \\
\midrule
Sparse DIB & 0.64 & 94 \\
RPEClust & 0.73 & 18193 \\
VarSelLCM & 0.14 & 0 \\
Sparse PCA/K-Means & 0.33 & 316 \\
PCA/K-Means & 0.23 & 18193\\
COSA/PAM & 0.50 & 18193\\
Sparse K-Means & 0.46 & 18193\\
\bottomrule
\end{tabular}
\label{tab:bladder_cancer_res}
\end{table}

The RPEClust algorithm yields the highest ARI with 0.73, followed by Sparse DIB and COSA/PAM. It is important to point out that the clustering problem on the bladder cancer data set is more of a subspace clustering problem, with clusters being defined on different subsets of features; this justifies the relative success of COSA. The main strengths of Sparse DIB are that it produces the second highest ARI score and also selects a subset of features, unlike the top-performing RPEClust. This enables an interpretation of the obtained partition. Sparse K-Means fails to select a subset of features with a higher ARI compared to when all variables are included, whereas VarSelLCM seems to identify all variables as redundant and returns a random partition.
\begin{figure}[!ht]
    \centering
\includegraphics[width=0.75\linewidth]{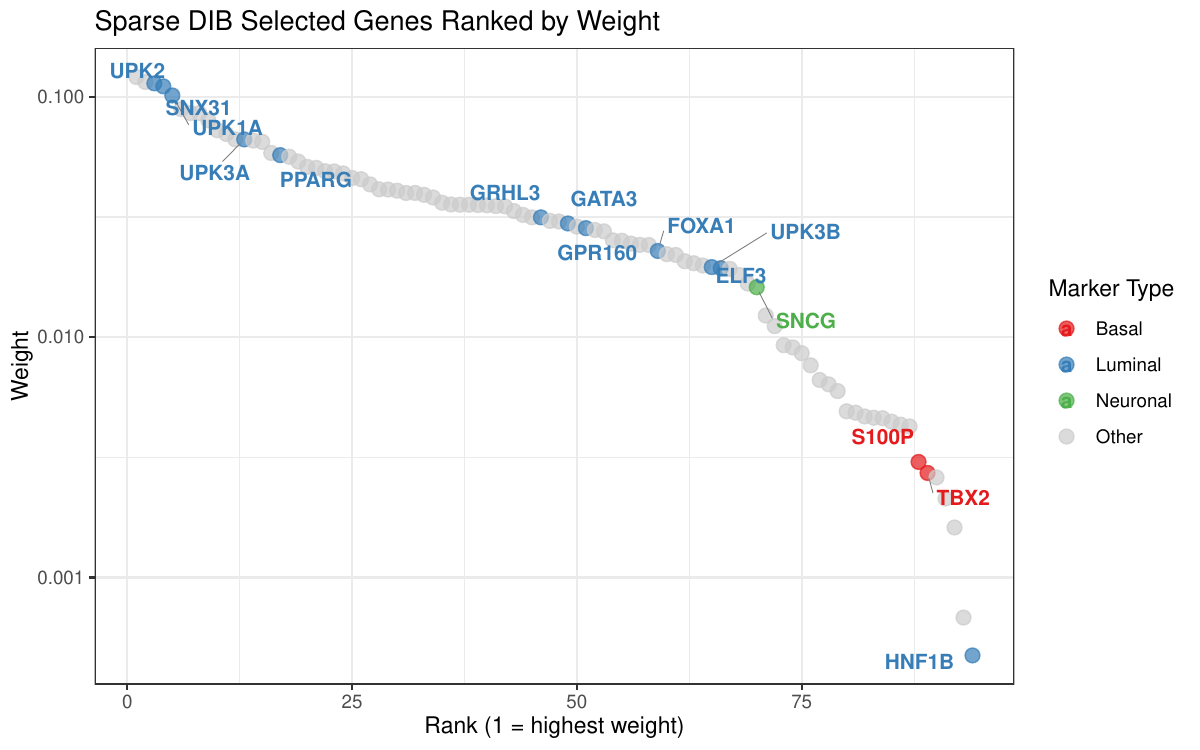}
    \caption{Genes with non-zero weights selected by Sparse DIB marked by type.}
    \label{fig:sdib_bladdercancer_weights}
\end{figure}
Figure \ref{fig:sdib_bladdercancer_weights} illustrates the 94 genes selected by Sparse DIB using a sparsity constraint value of $u = 3.3$. Of these 94 genes, twelve are known luminal markers, two are basal markers, and one is a neuronal marker, with the remaining 79 genes representing novel discriminative features or genes involved in supporting biological processes \cite{robertson2017comprehensive}. The weight distribution reflects the optimisation objective; Sparse DIB prioritises the largest and most heterogeneous class, namely luminal subtypes (approximately 60\% of the samples) by assigning substantially higher weights to luminal-discriminative features. The four uroplakins (UPK1A, UPK2, UPK3A, UPK3B), which are bladder-specific urothelial differentiation markers, collectively account for nearly 40\% of the total weight budget (combined weight 0.395), with UPK2 ranking third overall. Additional high-weight luminal markers include key transcription factors (GATA3, FOXA1, GRHL3, ELF3) and PPARG, a central regulator of luminal differentiation. In contrast, basal markers (S100P, TBX2) and the neuronal marker (SNCG) receive substantially lower weights which is still sufficient for discriminating these smaller and more molecularly homogeneous classes. This weight allocation is consistent with the information-theoretic objective of the algorithm, features that reduce uncertainty about the dominant luminal class contribute more to overall clustering quality. Notably, the canonical luminal marker KRT20 distinguishes luminal-papillary from other luminal subtypes and was not selected, demonstrating that the method correctly avoids features that introduce within-class heterogeneity in the aggregated three-class task.

\section{Conclusion}\label{sec:conclusion}

This paper introduced Sparse DIB, an extension of the DIB clustering framework to high-dimensional sparse data. Our simulation study demonstrated that Sparse DIB is a competitive alternative to high-dimensional clustering methods, including Sparse K-Means and latent class modelling with variable selection. This competitiveness was further confirmed on a real-world genomics data set of bladder cancer patients, where Sparse DIB performed reasonably well in differentiating between cancer subtypes, while selecting a small and interpretable subset of features. Notably, several of the selected genes have known clinical relevance, underscoring the practical utility of our approach.

Despite these promising results, the use of mutual information for simultaneous clustering and feature weighting warrants deeper theoretical investigation. Future work includes extending the proposed framework to enable sparse hierarchical agglomerative clustering, building on information-theoretic merging criteria \cite{slonim1999agglomerative}. Additionally, developing a version that incorporates cluster-specific feature weights would allow structures to be defined on different feature subsets, thus enhancing flexibility. Finally, a promising direction is to develop an extension for high-dimensional mixed-type data (e.g. combining genetic information with clinical variables), thus creating a unified tool for extracting coherent clusters from complex datasets with heterogeneous features.

\bibliographystyle{bibtex/splncs03_unsrt}
\bibliography{references}
\end{document}